\documentclass{article}

\usepackage{PRIMEarxiv}

\usepackage[utf8]{inputenc} 
\usepackage[T1]{fontenc}    
\usepackage{hyperref}       
\usepackage{url}            
\usepackage{booktabs}       
\usepackage{amsfonts}       
\usepackage{nicefrac}       
\usepackage{microtype}      
\usepackage{lipsum}
\usepackage{fancyhdr}       
\usepackage{graphicx}       
\usepackage[toc,page]{appendix}
\usepackage[autostyle]{csquotes}  
\graphicspath{{media/}}     

\usepackage{threeparttable}
\usepackage{tabularx}
\usepackage{bm}
\usepackage{graphicx}

\usepackage[backend=biber, style=numeric]{biblatex}
\title{Fine-Tuned ‘Small’ LLMs (Still) Significantly \\ Outperform Zero-Shot Generative AI \\ Models  in Text Classification}

\author{
  Martin Juan José Bucher \\
  Stanford University \\
  \texttt{mnbucher@stanford.edu} \\
   \And
  Marco Martini \\
  University of Zurich \\
  \texttt{marco.martini@uzh.ch} \\
}

\begin{document}
\maketitle

\begin{abstract}

Generative AI offers a simple, prompt-based alternative to fine-tuning smaller BERT-style LLMs for text classification tasks. This promises to eliminate the need for manually labeled training data and task-specific model training. However, it remains an open question whether tools like ChatGPT can deliver on this promise. In this paper, we show that smaller, fine-tuned LLMs (still) consistently and significantly outperform larger, zero-shot prompted models in text classification. We compare three major generative AI models (ChatGPT with GPT-3.5/GPT-4 and Claude Opus) with several fine-tuned LLMs across a diverse set of classification tasks (sentiment, approval/disapproval, emotions, party positions) and text categories (news, tweets, speeches). We find that fine-tuning with application-specific training data achieves superior performance in all cases. To make this approach more accessible to a broader audience, we provide an easy-to-use toolkit alongside this paper. Our toolkit, accompanied by non-technical step-by-step guidance, enables users to select and fine-tune BERT-like LLMs for any classification task with minimal technical and computational effort.

\end{abstract}


\section{Introduction}

The last decade has seen a paradigm shift in Natural Language Processing (NLP) with the advent of pre-trained Large Language Models (LLMs). These models have not only achieved previously unseen performance on a wide range of benchmarks. They have also shifted expectations for the future of artificial intelligence (AI) more generally.

Traditionally, leveraging LLMs involves two sequential steps: First, models are pre-trained on large text corpora to instill general language understanding. Pre-training is typically performed by specialized experts who then make the pre-trained models publicly available by providing all parameters and model states via checkpoints. Subsequently, these models can be further fine-tuned for specific tasks using smaller, dedicated training datasets, a task usually carried out by end-users who need to implement deep learning code and provide application-specific training data in the process. LLMs of this kind, like BERT or RoBERTa, have since been the state-of-the-art in numerous NLP tasks. Empirical evidence has shown that fine-tuned LLMs outperform traditional methods such as dictionaries, bag-of-words models, or approaches based on word embeddings \cite{devlin2018bert, liu2019roberta, yang2019xlnet, clark2020electra}. 

More recently, with models continuously growing in size and complexity, instruction-tuned generative AI models such as ChatGPT and Claude Opus have emerged. These models are not only significantly larger, but also more versatile than BERT-style models: While generative AI models are also pre-trained, they can be directly prompted via text commands to perform tasks without the need for a further fine-tuning step. As a result, generative AI promises a straightforward and intuitive user interaction while challenging the traditional approach, which requires fitting a model to labeled data. Although the empirical picture is still far from decisive, first studies suggest that generative AI models are already outperforming fine-tuned LLMs in text classification tasks \cite{gilardi2023chatgpt, zhong2023can}. 

In this paper, we contribute to the ongoing debate by systematically comparing the text classification performance of three prompt-instructed generative AI models (ChatGPT-3.5, ChatGPT-4, and Claude Opus) as well as Facebook's BART with that of several smaller, fine-tuned LLMs.

Our analysis spans a wide range of text classification tasks, including sentiment analysis, approval/disapproval recognition, emotion detection, and identification of political party positions. We also evaluate model performance across different text categories, such as news articles, tweets, and political speeches. Specifically, we conduct four case studies: (i) sentiment analysis of The New York Times coverage on the U.S. economy, (ii) stance classification of tweets about Brett Kavanaugh’s nomination to the U.S. Supreme Court, (iii) emotion detection in German political texts, and (iv) multi-class stance classification of nationalist party positions on European integration.

Our results demonstrate that fine-tuning smaller BERT-style models significantly outperforms generative AI models such as ChatGPT and Claude Opus (used in a "zero-shot" fashion) across all four applications when moderate amounts of training data for fine-tuning are provided. This tendency is especially pronounced for more specialized, non-standard classification tasks. Overall, these findings suggest that smaller, fine-tuned LLMs (still) constitute the state-of-the-art in text classification. 

Our results also imply that model size and complexity are no sufficient substitute for application-specific training data. For text classification tasks, the lightweight, tailored approach of fine-tuning small LLMs via training data remains preferable to the heavy-duty, one-size-fits-all approach of zero-shot prompting generative AI models.

Given our clear-cut results in favor of fine-tuning, we make available an easy-to-use toolkit with this paper. Presented as a simple Jupyter Notebook built on top of Hugging Face, our toolkit simplifies the process of fine-tuning LLMs, which typically requires deep learning and programming experience. It allows users to select and fine-tune smaller pre-trained LLMs for any classification problem (e.g., sentiment) or text category (e.g., news) with minimal requirements.

Our toolkit supports different languages and handles both binary and non-binary classification problems. It includes pre-implemented methods to address class imbalance, common in many applications (e.g., news texts typically have more negative than positive sentiment). While the notebook allows for the configuration and optimization of deep-learning hyperparameters, it comes with default hyperparameters that deliver strong performance out-of-the-box. Computationally intensive hyperparameter tuning is therefore an option but not required. The modular design of the toolkit allows for the integration of additional/future model releases. We further provide a detailed documentation and step-by-step user guidance. 

To further facilitate use and adoption of fine-tuning, we provide a non-technical introduction to the functioning principles of LLMs, clarify how these models expand on earlier text-as-data methods, and explain why this results in significant performance gains (Section 3). We also discuss current research areas, potential future trends in NLP and AI, and the opportunities of few-shot learning (Section 6).

\section{Related Work and Contribution}

A growing literature evaluates and compares the performance of fine-tuned and prompt-based LLMs \cite{bosley2023we, brown2020language, chae2023large, chen2021better, wei2021finetuned, edwards2024language, gilardi2023chatgpt, zhong2021adapting, zhong2023can}. However, some of this work does not specifically address text classification, which is the primary focus of our paper, but instead concentrates on various other text comprehension tasks \cite{brown2020language, wei2021finetuned}. Other studies primarily compare the performance of different fine-tuning methods \cite{chen2021better}, or evaluate various zero-shot approaches against each other \cite{zhong2021adapting}. 

The existing studies that come closest to our work, as they compare fine-tuned and prompt-based models for text classification tasks, are \cite{bosley2023we} and \cite{edwards2024language} as well as \cite{zhong2023can} and \cite{gilardi2023chatgpt}. However, these papers provide conflicting evidence as to which approach is superior for text classification.

For example, \cite{edwards2024language} compare a fine-tuned RoBERTa model with prompt-based models like GPT-3.5 and Meta’s Llama models across different classification tasks. The authors find that fine-tuned models (trained on the entire dataset) generally outperform prompt-based models. Similarly, \cite{bosley2023we} compare the performance of RoBERTa and GPT-3 in a study on an English-language dataset of parliamentary speeches, concluding that fine-tuning yields better results. 

In contrast, recent work by \cite{zhong2023can} and \cite{gilardi2023chatgpt} suggests that ChatGPT has caught up with or even surpassed fine-tuned models. \cite{zhong2023can} find that ChatGPT is on par with BERT-style models for some text understanding tasks, though results are mixed overall. \cite{gilardi2023chatgpt} present results suggesting that ChatGPT even outperforms human annotators for text labeling tasks. 

Inspired by these works, we aim to provide an up-to-date, systematic empirical analysis that offers a comprehensive view on the performance of fine-tuned versus prompt-based models for text classification. 

While our results clearly point in the same direction as the studies by \cite{bosley2023we} and \cite{edwards2024language}, our paper expands on the existing literature in several ways:

\begin{itemize}

    \item \textit{Comprehensive Comparison}: We systematically compare model performance across a diverse set of classification tasks and text categories. This allows us to identify performance variations across tasks and highlight areas where the differences between fine-tuned and prompt-based models are most (and least) pronounced.
    \item \textit{Latest Generative AI Models}: We provide results for the latest generation of prompt-based generative AI models (GPT-4 and Claude Opus). This is crucial, as previous studies based on older models like GPT-3 offer limited insights into the current performance balance.
    \item \textit{Nuanced Understanding of Fine-Tuned LLMs}: By investigating the performance of several fine-tuned LLMs, we offer a detailed understanding of how these models compare across classification tasks. This helps identify which models are better suited for specific tasks, thereby aiding user choice. Such a comparison provides a broader perspective on the current LLM landscape for text classification.
    \item \textit{Ablation Studies and Training Data Impact}: We conduct ablation studies and analyze the effect of training data size on model performance for fine-tuning. This offers valuable insights into the number of labeled samples required to achieve optimal performance, a crucial consideration given our consistent findings in favor of fine-tuned LLMs.
    \item \textit{Accessible Toolkit for Fine-Tuning}: We offer a one-stop-shop toolkit for text classification to enable a wider audience to use pre-trained LLMs for text classification tasks.

\end{itemize}

With these contributions, we aim to provide a comprehensive and practical resource for understanding and applying LLMs in text classification.

\section{Non-technical Background: From Keywords to Large Language Models}

This section provides a non-technical introduction to LLMs, which complements our empirical results and our provided fine-tuning toolkit. It explains the functioning principles of LLMs and the reasons for their superior performance compared to earlier methods. To provide a comprehensive overview, we first briefly cover traditional hand-coding and dictionary approaches before discussing machine learning methods and LLMs.

Hand-coding is among the earliest text-as-data approaches. When hand-coding, \textit{Human} coders classify verbal information according to pre-specified codebooks and explicit coding rules \cite{handc1}, \cite{handc2}. Because hand-coding leverages human text-understanding, it is relatively easy to perform and yields high-quality results. However, hand-coding is labor-intensive — both regarding codebook development and the coding process itself. This makes it expensive and slow (see Figure 1). Moreover, hand-coding can be inconsistent across coders and within coders over time, making it partially non-replicable. High costs, slow procedures, and limited reproducibility are thus clear drawbacks of hand-coding approaches.

Dictionary approaches alleviate some of these concerns. These methods automatically extract information from text by mapping a larger set of keywords (or keyword combinations) to a small set of categories of interest. For example, sentiment dictionaries classify texts according to their content of positive and negative words or phrases \cite{newssenti, silva2021twittersenti}. Dictionaries can be fully automated and therefore fast, replicable, and transparent. However, because dictionaries only take selected keywords into account, they ignore most nuanced textual information. Because words have different meanings in different settings, dictionaries can also be context-dependent and noisy when used off-the-shelf. Moreover, dictionary construction in itself is a difficult and time-consuming process and requires iterative testing to minimize false positives and false negatives. Thus, while dictionaries have advantages in terms of speed and replicability, they remain a relatively blunt tool.

\begin{figure*}
    \centering
    \begin{minipage}[t]{1.0\textwidth}
  	  \begin{center}
  	  \vspace*{.3cm}
  	  \includegraphics[width=.98\textwidth]{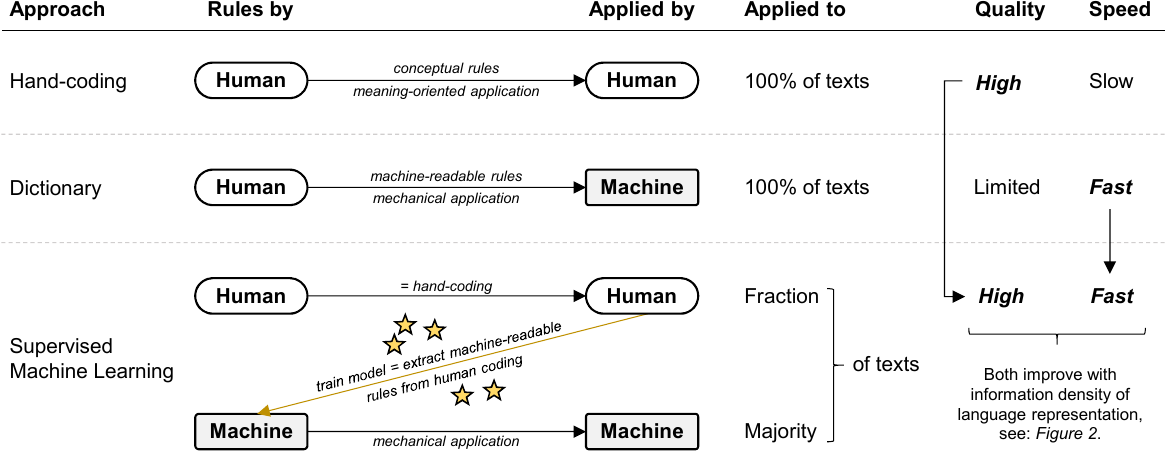}   
  	  \vspace*{.3cm}
  	  \end{center}
    \end{minipage}
    \caption{Overview of existing text-as-data methods and their characteristics: Machine learning approaches for text classification have the \emph{potential} to combine the advantages of hand-coding (high-quality) and dictionaries (speed), while avoiding their respective downsides. The degree to which this potential can be realized in practice depends on the underlying text representation (see Figure 2).}
    \label{fig:method-overview}
\end{figure*}

Machine learning methods follow a different approach. Unlike hand-coding and dictionary approaches, which both rely on \textit{explicit} human-crafted rules, machine learning techniques derive \textit{implicit} rules from human-labeled data. In supervised learning, a model extracts classification-relevant patterns from a sample of manually-coded (labeled) training data. These patterns are then used to auto-classify the bulk of the remaining data \cite{miller2020active, erlich2022multilabel, chang2020lstm}. Machine learning methods allow human coders to classify texts based on natural human language. This frees coders from having to create abstract rules and instead allows them to follow their intuitive language understanding. Humans can thus focus on the information content of texts, while the learning algorithm extracts predictive regularities from the relation between text features and labels.

The combination of human-labeled data and automated rule extraction is powerful because it can fuse the high quality of human coding with the speed of automated approaches (see Figure 1). This combination works best, however, with more sophisticated and information-dense numerical language representations. Deep-learning approaches based on LLMs are now making it possible to realize this potential. To see this, we next describe the evolution of language representations from bag-of-words models via word embeddings to LLMs and point out why each step has generated significant performance increases in NLP tasks.

\subsection{Bag-Of-Words}

Early machine learning approaches for automated text analysis rest on \textit{bag-of-words} models \cite{bow3}. These models treat text documents as collections (or bags) of words without regard to word order and grammar. This allows representing complex language information in a simple spreadsheet format that is easy to work with in downstream applications. In this format, rows represent documents, columns represent words, and cells record how often a word occurs in a document. Thus, a corpus with $D$ documents and $W$ unique words is represented as a $D$x$W$ matrix. Each row (corresponding to a document) then captures the word signature (i.e., word distribution) of the original input text.

While bag-of-word approaches can be powerful, they have notable shortcomings. First, they tend to produce large sparse matrices - because most words typically do not occur in most documents. This also results in matrices with more columns (words = variables) than rows (documents = observations), leading to $P>N$ problems that are unwieldy in regression contexts.

Second, the spreadsheet representation treats all words as equally different from each other. For example, bag-of-words representations cannot capture that the words 'cake' and 'cookie' have more in common than the words ‘cake’ and ‘court’. Bag-of-words approaches are oblivious to the meaning of words and cannot reflect relations between words.

Third, bag-of-words approaches discard all information contained in word order, sentence structure, and grammar. This ignores the entire narrative content of a text and makes it difficult to retain information beyond the general topic of a document. This problem is further aggravated by pre-processing steps such as stopword removal, which typically discards negations, and thus discards further potentially important information.

\subsection{Word Embeddings}

In contrast to bag-of-words models, word embeddings capture core aspects of word meaning and similarity \cite{wordemb}. Word embeddings represent words as vectors in a high-dimensional space, where one or multiple dimensions can be thought of as corresponding to a particular characteristic. For example, one dimension might represent the degree of positivity of a word, while another might represent its length or its frequency of use. Consequently, similar words are (ideally) mapped to vectors that point to a similar location in vector space.\footnote{Word embeddings are created by training neural networks to learn co-occurrence patterns of words. Well-known word embedding models are Word2Vec \cite{mikolov2013efficient}, GloVe \cite{pennington2014glove}, and FastText \cite{bojanowski2017enriching}.}

Word embeddings are a richer (and usually more compact) numerical representation of natural language than bags-of-words matrices. This has several advantages. First, word embeddings reduce the dimensionality of the language representation, which solves the $P>N$ issue and facilitates the use of secondary models for downstream tasks.

Second, word embeddings can extrapolate to words that do not occur in the training data for a downstream task. For example, when training a sentiment model where the training data contains positive references to 'apples' and 'bananas,' then the model will likely also attribute a positive sentiment to 'oranges' - which will be located close to other fruits in the embedding space.

Third, word embeddings can reflect further application-relevant intricacies of word meaning. For instance, embeddings can capture that 'man' is related to 'woman' and 'boy' is related to 'girl' on one dimension (gender), but also that 'man' is related to 'boy' and 'woman' is related to 'girl' on another dimension (age).

However, word embeddings still focus on words and continue to abstract from sentence structure and word order. They will thus only be able to capture the information content of longer segments of text to a limited extent.

\begin{figure*}[h!]
    \centering
    \begin{minipage}[t]{1.0\textwidth}
        \begin{center}
  	  \vspace*{.3cm}
  	  \includegraphics[width=.98\textwidth]{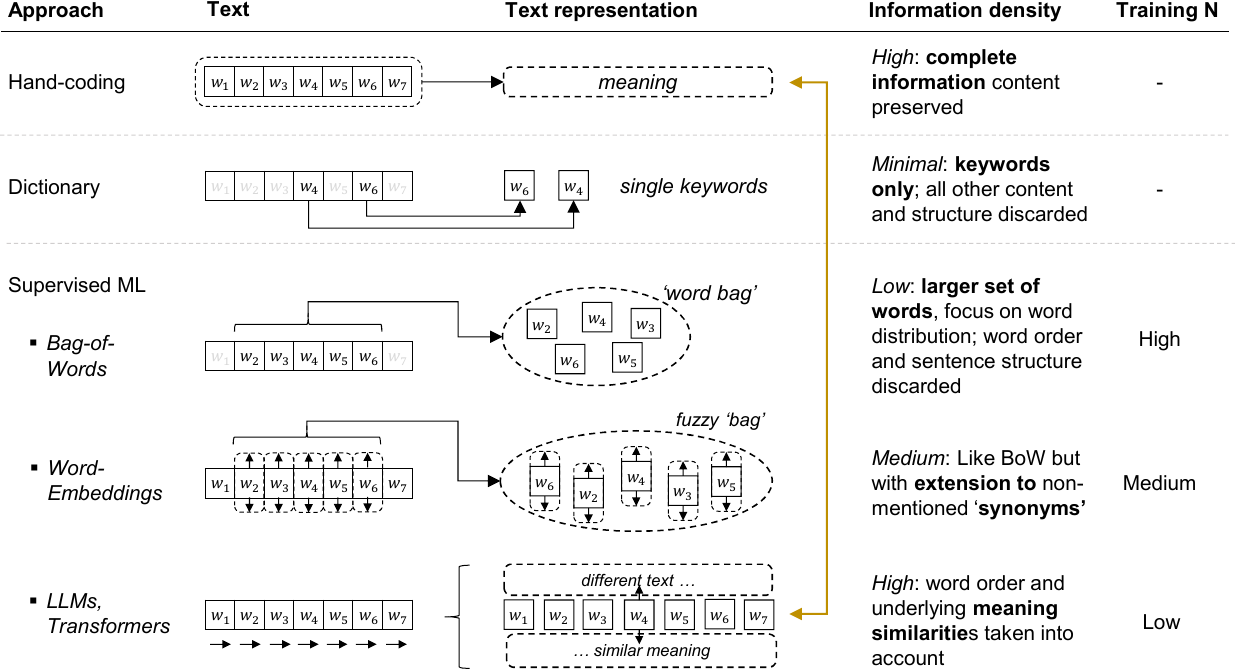}
  	  \vspace*{.3cm}
  	  \end{center}
    \end{minipage}
    \caption{Text representation of different text-as-data approaches: Existing approaches differ starkly in the sophistication of their text representation. Pre-trained LLMs approximate a more holistic human text understanding that focuses on meaning (\emph{concepts} rather than form (\emph{wording}. This allows LLMs to effectively leverage the information contained in text. By contrast, earlier text-as-data approaches discard significant information due to their more rudimentary language representations.}
    \label{fig:method-overview}
\end{figure*}

\subsection{Pre-trained Large Language Models}

Pre-trained LLMs derive their name from the fact that, before being fine-tuned or prompted to perform a \textit{specific} task, they are pre-trained on large amounts of text to 'learn' the \textit{general} structure and logic of language. The pre-training process encodes a language representation that focuses on sequences of text rather than individual words. For BERT-style (encoder-only) models, fine-tuning later enriches this language representation with task-specific information. For GPT-style (decoder-only) models, prompts induce an answer generation process that is, in essence, a text prediction exercise resting on the model's pre-trained language understanding.

Pre-training relies on \emph{unsupervised} learning. Unlike in supervised learning, the models are not given explicit instructions via 'labeled' training data. Instead, they discover relationships by predicting (temporarily masked) words in their training texts based on the surrounding word sequences. Encoder-style models have typically been trained with a Masked Language Modeling (MLM) objective, where random words are 'masked out' and the model must predict the missing words. In contrast, decoder-style models have typically been trained with a Causal Language Modeling (CLM) objective, where the goal is to predict the next token (or word) in the sequence.

Pre-training is much more resource-intensive than fine-tuning: Learning the complexities of natural language requires long exposure to large volumes of text. Hence, although the pre-training phase is unsupervised and requires no manual input, it was long constrained by excessive computational costs. This changed with the development of the Transformer architecture \cite{vaswani2017attention}. Transformers can be more effectively parallelized than earlier sequence-learning approaches such as Recurrent Neural Networks (RNNs) and Long Short-Term Memory models (LSTMs). This paved the way for models such as BERT (Bidirectional Encoder Representations from Transformers; \cite{devlin2018bert} and much bigger models such as the GPT series (Generative Pre-trained Transformer; \cite{radford2018improving}.\footnote{While decoder-style models such as the GPT series are typically several orders of magnitude larger (in number of parameters) than encoder-style models, we refer to all of them as LLMs in this paper.}

Analogously to how word embeddings capture similarities between words, LLMs capture similarities between longer text sequences. For example, LLMs 'understand' that the sentences 'this is an article on natural language processing' and 'you are reading a publication on automated text analysis' are similar in meaning despite their different wording. LLMs also 'understand' that 'this is an article on natural language processing' and 'this is \emph{not} an article on natural language processing' differ in meaning despite similar wording. LLMs take negations and related structures into account and are able to correctly interpret the associated shifts in meaning due to their attention mechanism in the underlying transformer architecture.

This explains the strong performance of pre-trained Transformer models on many downstream tasks. Equivalent to how 'apples' and 'bananas' are represented in a similar part of the vector space of word embeddings, word sequences with similar meanings are represented as being similar within the internal representation of an LLM. This allows the models to effectively extrapolate to unseen text sequences once they have been fine-tuned or prompted to perform a downstream task (also see \cite{laurer2023less}.

For BERT-style encoder models, fine-tuning then constitutes a secondary training step that complements the unsupervised pre-training stage with a \textit{supervised} task-orientation stage. It involves feeding the model with a small set of labeled data. During fine-tuning, the model thus learns the nuanced relationships between text sequences and the labels associated with them. This 'knowledge' can then be used to classify unseen text sequences by predicting their most likely label.\footnote{Except for the more nuanced language representation contained within the pre-trained model, fine-tuning is similar to the use of supervised learning with bag-of-words or word embeddings representations.}

For GPT-style decoder models, fine-tuning is not \emph{per se} required. However, model performance following prompts will generate high-quality output only if the model has seen sufficient examples of the task at hand during its pre-training (\& post-training) phase. As we discuss in more detail in Section 6, we argue that for most classification tasks this requirement appears to not be sufficiently satisfied — thereby explaining the better performance of fine-tuned models in our analysis.

\section{Method: Fine-tuned LLMs vs. Zero-Shot Generative AI Models}

In the following, we compare the text classification performance of smaller, fine-tuned LLMs with that of three major generative AI models (ChatGPT with GPT-3.5 / GPT-4, and Claude Opus) and BART. We evaluate all models across four heterogeneous case studies, which span a diverse set of classification tasks. These tasks vary by concept (sentiment, stance, and emotions), text form (news, tweets, and speeches), language (English and German), and number of classes (binary vs. multi-class).

For smaller, fine-tuned LLMs, we assess five leading models of different sizes and architectures (RoBERTa Base, RoBERTa Large, DeBERTa V3, Electra Large, and XLNet). We fine-tune these models using our toolkit provided with this paper.\footnote{Our toolkit, including detailed step-by-step guidance, is available \href{https://github.com/mnbucher/text-cls-llms}{here}.} We use our default hyperparameter setup throughout to showcase the capabilities of this setup, which we argue will be of greatest interest to most users, as no hyperparameter tuning is required (but possible; see Figure \ref{fig:method-overview} for a suggested workflow summary).

\begin{figure*}[h!]
    \centering
    \begin{minipage}[t]{1.0\textwidth}   
  	   \begin{center}
  	  \vspace*{.2cm}
        \includegraphics[width=.97\textwidth]{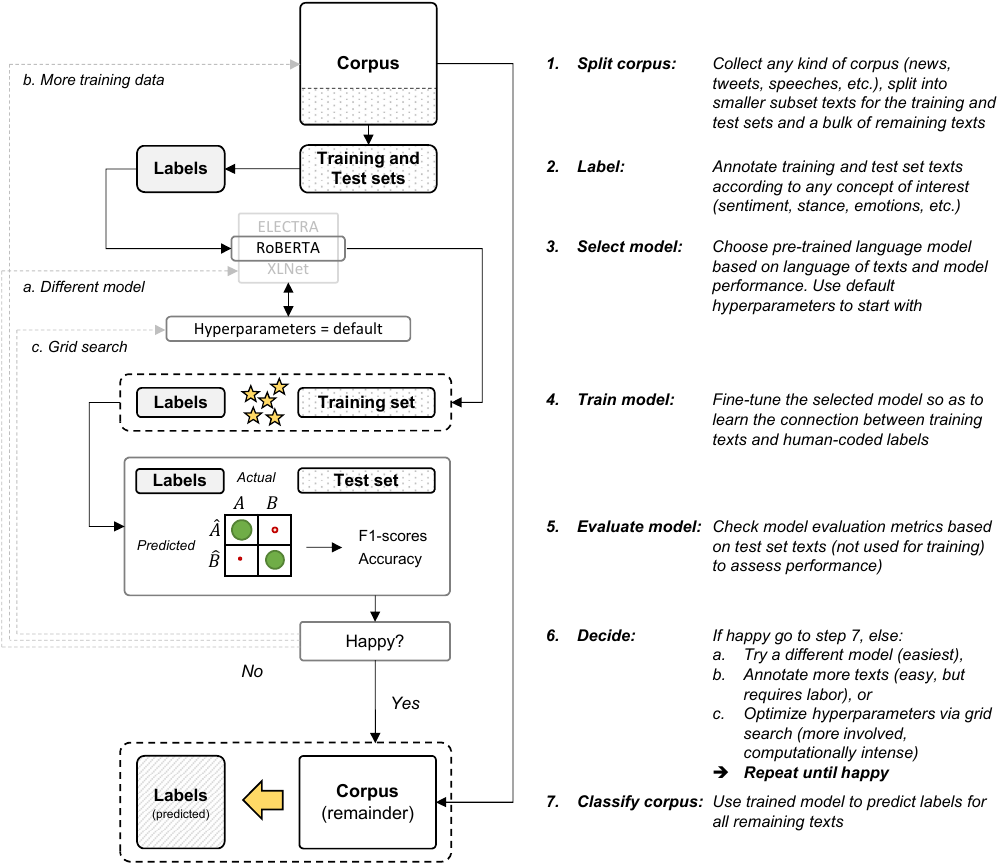}
  	  \vspace*{.2cm}
  	  \end{center}
    \end{minipage}
    \caption{Schematic representation of the workflow with our toolkit}
    \label{fig:method-overview}
\end{figure*}

For all fine-tuned models, we proceed in the same manner across all case studies. Each dataset is split into training and test sets, with the test set reserved for evaluating model predictions on unseen data. We then fine-tune the models for each case study and compare their classification performance on the test set using standard metrics: Accuracy, Precision, Recall, and F1-scores. To account for variability due to random seed initialization, we train each model three times with different seeds and report the mean and standard deviation for each metric across the three runs.

For the zero-shot LLMs, we prompt all models to label the full datasets we used for evaluating the fine-tuned models (and not just the test sets). The prompts include a description of the dataset and the classification categories as well as the instruction to assign the individual observations to the most appropriate category (detailed prompts for all models are provided in the online appendix). Because the zero-shot approach requires no training data for fine-tuning, we do not split the data into training and test sets. Instead, we prompt the models to label all observations in the dataset. We then evaluate model performance by comparing the model-generated labels with the original human-annotated labels, using the same performance metrics as for the fine-tuned models. We provide the full prompts in Appendix A.

In summary, we evaluate the following models across our case studies:

\begin{itemize}
    \item \textbf{MAJ-VOT} — As a baseline, we use a \textit{failed classifier} that assigns the majority class to all observations. For imbalanced datasets, this can suggest good performance on metrics sensitive to class-imbalance (e.g., Accuracy). However, metrics such as Recall or F1-Score reveal that this classifier performs poorly for the minority class(es).
    \item \textbf{ROB-BASE} (125M) — RoBERTa Base \cite{liu2019roberta}, provided as \texttt{'roberta-base'} via Hugging Face (HF) \cite{wolf2020transformers}, is pre-trained in a self-supervised manner using masked language modeling (MLM).
    \item \textbf{ROB-LRG} (355M) — RoBERTa Large, provided as \texttt{'roberta-large'}, is similar to RoBERTa Base but has more parameters. This boosts performance but increases computational costs.
    \item \textbf{DEB-V3} (435M) — DeBERTa, provided as \texttt{'microsoft/deberta-v3-large'}, builds on BERT/RoBERTa. Despite its similar size, it takes substantially longer to fine-tune because of differences in architecture.
	\item \textbf{ELE-LRG} (335M) — ELECTRA Large, provided as \texttt{'google/electra-large-discriminator'} and \texttt{'german-nlp-group/electra-base-german-uncased'}, was introduced by \cite{clark2020electra}. ELECTRA claims to be more compute-efficient than existing alternatives.
    \item \textbf{XLNET-LRG} (340M) — XLNet \cite{yang2019xlnet}, provided as \texttt{'xlnet-large-cased'}, was pre-trained with an autoregressive method to learn bidirectional contexts and has beens shown to perform comparable to RoBERTa.
    \item \textbf{BART-LRG} (407M) — BART was introduced by Facebook \cite{lewis2019bart} and uses an encoder-decoder model. BART can generate text similar to ChatGPT and Claude, although relying on a different architecture. More details are provided in Appendix B.
    \item \textbf{GPT-3.5} (175B+) / \textbf{GPT-4} (n/a) — ChatGPT, developed by OpenAI \cite{ouyang2022training}. It has substantially more parameters than any of the previsouly listed models (GPT-3.5 = 175B, GPT-4 $\approx$ 1.8T \footnote{The exact number of parameters for GPT-4 is unknown given OpenAI no longer reveals technical details of its models but many sources claim it might be around 1.8T}. ChatGPT can process complex input texts and generate text of varying length as output. We use the \texttt{'gpt-3.5-turbo'} and \texttt{'gpt-4-1106-preview'} checkpoints provided by their API. Details about prompting are provided in Appendix A.
    \item \textbf{CLD-OPUS} (n/a) — Claude 3 Opus, developed by Anthropic \cite{anthropic2024claude}. The model size and technical details remain undisclosed. Opus can handle complex input prompts on par with GPT4, according to various benchmarks and empirical results published by Anthropic. We use the \texttt{'claude-3-opus-20240229'} checkpoint provided by their API. Details about prompting are provided in Appendix A.
\end{itemize}

\section{Results}

This section presents the results of our comparison. The first four subsections detail the findings for each case study for both fine-tuned and zero-shot prompted models. The fifth subsection reports the results of our ablation studies, which analyze the impact of training data size on the performance of our preferred fine-tuned model, RoBERTa Large.

\subsection{Sentiment Analysis on The New York Times Coverage of the US Economy}

In our first case study, we perform \textit{sentiment analysis} — perhaps the most widely-studied problem in NLP. Sentiment analysis involves classifying statements into (i) positive and negative or (ii) positive, negative, and neutral categories.

The dataset is taken from \cite{barbera2021automated}. It contains information on the sentiment of articles about the US economy from The New York Times. In the original study, the authors compare three dictionary-based approaches for sentiment analysis with a supervised machine learning approach (bag-of-words in combination with logistic regression). The machine learning model is reported to achieve an Accuracy of 0.71 and a Precision of 0.713, clearly surpassing the dictionaries (highest Accuracy = 60.5, highest Precision = 45.7).

\begingroup 
\renewcommand{\arraystretch}{1.2} 

\begin{table*}[h!]
\centering
\begin{threeparttable}
\caption{Results for Sentiment Analysis (US Economy)}
\label{tab:exps-sentiment-analysis}
\begin{tabular}{l l l l l l}
\hline
\textbf{Model Name} & \textbf{Accuracy} & \textbf{Prec. (wgt.)} & \textbf{Recall (wgt.)} & \textbf{F1 (macro)} & \textbf{F1 (wgt.)} \\
\hline
MAJ-VOT & $0.73\,\scriptstyle{(\pm0.00)}$ & $0.53\,\scriptstyle{(\pm0.00)}$ & $0.73\,\scriptstyle{(\pm0.00)}$ & $0.42\,\scriptstyle{(\pm0.00)}$ & $0.61\,\scriptstyle{(\pm0.00)}$ \\
\midrule
ROB-BASE & $0.89\scriptstyle{\,(\pm0.00)}$ & $0.89\,\scriptstyle{(\pm0.01)}$ & $0.89\,\scriptstyle{(\pm0.00)}$ & $0.86\,\scriptstyle{(\pm0.01)}$ & $0.89\,\scriptstyle{(\pm0.01)}$ \\
\textbf{ROB-LRG} & $\bm{0.92\,\scriptstyle{(\pm0.01)}}$ & $\bm{0.92\,\scriptstyle{(\pm0.01)}}$ & $\bm{0.92\,\scriptstyle{(\pm0.01)}}$ & $\bm{0.90\,\scriptstyle{(\pm0.01)}}$ & $\bm{0.92\,\scriptstyle{(\pm0.01)}}$ \\
DEB-V3 & $0.92\,\scriptstyle{(\pm0.02)}$ & $0.92\,\scriptstyle{(\pm0.01)}$ & $0.92\,\scriptstyle{(\pm0.02)}$ & $0.90\,\scriptstyle{(\pm0.02)}$ & $0.92\,\scriptstyle{(\pm0.01)}$\\
ELE-LRG & $0.90\,\scriptstyle{(\pm0.01)}$ & $0.90\,\scriptstyle{(\pm0.01)}$ & $0.90\,\scriptstyle{(\pm0.01)}$ & $0.88\,\scriptstyle{(\pm0.02)}$ & $0.90\,\scriptstyle{(\pm0.01)}$ \\
XLNET-LRG & $0.81\,\scriptstyle{(\pm0.01)}$ & $0.85\,\scriptstyle{(\pm0.01)}$ & $0.81\,\scriptstyle{(\pm0.01)}$ & $0.78\,\scriptstyle{(\pm0.01)}$ & $0.82\,\scriptstyle{(\pm0.01)}$ \\
\hline
BART-LRG & $0.85\,\scriptstyle{(\pm0.00)}$ & $0.84\,\scriptstyle{(\pm0.00)}$ & $0.85\,\scriptstyle{(\pm0.00)}$ & $0.80\,\scriptstyle{(\pm0.00)}$ & $0.84\,\scriptstyle{(\pm0.00)}$ \\
GPT-3.5  & $0.82\,\scriptstyle{(\pm0.00)}$ & $0.84\,\scriptstyle{(\pm0.00)}$ & $0.82\,\scriptstyle{(\pm0.00)}$ & $0.79\,\scriptstyle{(\pm0.00)}$ & $0.83\,\scriptstyle{(\pm0.00)}$ \\
GPT-4 & $0.87\,\scriptstyle{(\pm0.00)}$ & $0.87\,\scriptstyle{(\pm0.00)}$ & $0.87\,\scriptstyle{(\pm0.00)}$ & $0.84\,\scriptstyle{(\pm0.00)}$ & $0.87\,\scriptstyle{(\pm0.00)}$ \\
CLD-OPUS & $0.86\,\scriptstyle{(\pm0.00)}$ & $0.87\,\scriptstyle{(\pm0.00)}$ & $0.86\,\scriptstyle{(\pm0.00)}$ & $0.83\,\scriptstyle{(\pm0.00)}$ & $0.87\,\scriptstyle{(\pm0.00)}$ \\
\hline
\end{tabular}
\begin{tablenotes}[para,flushleft] \small
\textit{Note}: Results for fine-tuned models on unseen test set with $N=200$. Results for BART, GPTs, and Claude on full data. Fine-tuned models use gradient accumulation with 8 steps and batch size 4, except DEB-V3 (batch size 2). 
\end{tablenotes}
 \end{threeparttable}
\end{table*}

\endgroup

For our analysis, we use the dataset '\texttt{3SU}'. It contains $4195$ paragraphs coded by undergraduates. The students used a 5-category coding scheme (negative, mixed, neutral, not sure, positive). We only use paragraphs coded as 'relevant' (i.e., US economy related). We map all paragraphs with a 'positive' label to $1$, all paragraphs with a 'negative' label to $0$ and ignore paragraphs from the other categories. We then aggregate all labels based on document ID with a majority voting scheme. The final dataset has $N=1374$ observations, where $N=200$ observations are set aside for the test set. The class distribution is relatively imbalanced with $374$ positive and $1000$ negative labels.

The task is to predict whether a paragraph about the US economy is positive or negative. We fine-tuned all models as described above. Results are presented in Table \ref{tab:exps-sentiment-analysis}. The strongest results are achieved by ROB-LRG and DEB-V3.\footnote{Weigthed metrics are scaled by class size so that larger classes receive more weight. Macro F1 is computed as the arithmetic (i.e., unweighted) mean of each F1 score by class.} Most other fine-tuned models except XLNET-LRG also achieve strong results with values around $0.90$ and small variances (indicating that the random seed has little effect on model performance). BART, ChatGPT, and Claude also perform well on the sentiment task but remain behind the fine-tuned models.

\subsection{Stance Classification on Tweets about Kavanaugh Nomination}

Our second case study investigates \textit{stance classification}. Stance and sentiment are conceptually different: Whereas sentiment reflects the tone of a text from positive to negative, stance captures a positional attitude from supportive to opposing. For example, a statement such as 'I am happy that the government was voted out of office.' has a \textit{positive} sentiment but an \textit{opposing} stance.

The dataset is taken from \cite{bestvater2022sentiment}\footnote{\texttt{kavanaugh\_tweets\_groundtruth.csv}}. It contains manually labeled tweets in which people express their view on the 2018 nomination of Brett Kavanaugh to the U.S. Supreme Court. In the original study, the authors evaluate two dictionary methods, a bag-of-words approach combined with a support-vector machine (SVM), and a BERT language model. Both machine learning approaches are reported to outperform the dictionaries. BERT achieves an F1-score of $0.938\,(\pm0.002)$, although the SVM performs almost identically.

\begingroup 
\renewcommand{\arraystretch}{1.2} 

\begin{table*}[h!]
\centering
\begin{threeparttable}
\caption{Results for Stance Classification (Nomination Approval)}
\label{tab:exps-stance-class}
\begin{tabular}{l l l l l l}
\hline
\textbf{Model Name} & \textbf{Accuracy} & \textbf{Prec. (wgt.)} & \textbf{Recall (wgt.)} & \textbf{F1 (macro)} & \textbf{F1 (wgt.)} \\
MAJ-VOT & $0.50\,\scriptstyle{(\pm0.00)}$ & $0.25\,\scriptstyle{(\pm0.00)}$ & $0.50\,\scriptstyle{(\pm0.00)}$ & $0.33\,\scriptstyle{(\pm0.00)}$ & $0.33\,\scriptstyle{(\pm0.00)}$ \\
ROB-BASE & $0.86\,\scriptstyle{(\pm0.01)}$ & $0.86\,\scriptstyle{(\pm0.01)}$ & $0.86\,\scriptstyle{(\pm0.01)}$ & $0.86\,\scriptstyle{(\pm0.01)}$ & $0.86\,\scriptstyle{(\pm0.01)}$ \\
ROB-LRG & $0.92\,\scriptstyle{(\pm0.01)}$ & $0.93\,\scriptstyle{(\pm0.01)}$ & $0.92\,\scriptstyle{(\pm0.01)}$ & $0.92\,\scriptstyle{(\pm0.01)}$ & $0.92\,\scriptstyle{(\pm0.01)}$ \\
\textbf{DEB-V3} & $\bm{0.94\,\scriptstyle{(\pm0.01)}}$ & $\bm{0.94\,\scriptstyle{(\pm0.01)}}$ & $\bm{0.94\,\scriptstyle{(\pm0.01)}}$ & $\bm{0.93\,\scriptstyle{(\pm0.01)}}$ & $\bm{0.94\,\scriptstyle{(\pm0.01)}}$ \\
ELE-LRG & $0.74\,\scriptstyle{(\pm0.01)}$ & $0.66\,\scriptstyle{(\pm0.02)}$ & $0.74\,\scriptstyle{(\pm0.01)}$ & $0.67\,\scriptstyle{(\pm0.02)}$ & $0.69\,\scriptstyle{(\pm0.02)}$ \\
XLNET-LRG & $0.83\,\scriptstyle{(\pm0.01)}$ & $0.83\,\scriptstyle{(\pm0.01)}$ & $0.83\,\scriptstyle{(\pm0.01)}$ & $0.83\,\scriptstyle{(\pm0.01)}$ & $0.83\,\scriptstyle{(\pm0.01)}$ \\
\hline
BART-LRG & $0.53\,\scriptstyle{(\pm0.00)}$ & $0.59\,\scriptstyle{(\pm0.00)}$ & $0.53\,\scriptstyle{(\pm0.00)}$ & $0.44\,\scriptstyle{(\pm0.00)}$ & $0.44\,\scriptstyle{(\pm0.00)}$ \\
GPT-3.5 & $0.53\,\scriptstyle{(\pm0.00)}$ & $0.58\,\scriptstyle{(\pm0.00)}$ & $0.53\,\scriptstyle{(\pm0.00)}$ & $0.48\,\scriptstyle{(\pm0.00)}$ & $0.47\,\scriptstyle{(\pm0.00)}$ \\
GPT-4 & $0.58\,\scriptstyle{(\pm0.00)}$ & $0.68\,\scriptstyle{(\pm0.00)}$ & $0.58\,\scriptstyle{(\pm0.00)}$ & $0.51\,\scriptstyle{(\pm0.00)}$ & $0.51\,\scriptstyle{(\pm0.00)}$ \\
CLD-OPUS & $0.61\,\scriptstyle{(\pm0.00)}$ & $0.68\,\scriptstyle{(\pm0.00)}$ & $0.61\,\scriptstyle{(\pm0.00)}$ & $0.57\,\scriptstyle{(\pm0.00)}$ & $0.57\,\scriptstyle{(\pm0.00)}$ \\
\hline
\end{tabular}
\begin{tablenotes}[para,flushleft] \small
\textit{Note}: Results for fine-tuned models on unseen test set with $N=200$. Results for BART, GPTs, and Claude on full data. Fine-tuned models use gradient accumulation with 8 steps and batch size 4, except DEB-V3 (batch size 2). 
\end{tablenotes}
 \end{threeparttable}
\end{table*}

\endgroup

We use the 'text' and 'stance' columns from the dataset and apply some pre-processing by removing 'RT' flags (Retweets), twitter handles, and URLs. Afterwards, several tweets have identical text content (original and cleaned re-tweets). We therefore aggregate identical tweets via majority voting on the class label. The final dataset has $N = 1173$ tweets with a balanced class distribution of $699$ observations for the $1$ label (supportive stance) and $674$ observations for the $0$ label (opposing stance). Again, we reserve $N=200$ observations for the test set.

The classification task is to predict whether a tweet supports or opposes Kavanaugh's nomination. We train the same models as before. Results are reported in Table \ref{tab:exps-stance-class}. We achieve the strongest results with DEB-V3 and ROB-LRG with a performance of $0.94$ and $0.92$ respectively across metrics. The other three fine-tuned models show some variation. However, all fine-tuned models surpass the prompt-based approaches: BART, ChatGPT, and Claude perform rather weak on this more subtle task and score only marginally above the naive majority vote baseline.

\subsection{Emotion Detection on Political Texts in German}

Next, we turn to \textit{emotion classification}, an application that has recently seen increased attention by social scientists \cite{gennaro2022emoreason, vish2022twitteremo, onsa2021speechemo}. Emotions affect political discourse, behavior, and opinion formation. Yet emotional tone goes beyond positive vs. negative sentiment or supportive vs. opposing stance. For example, while anger and fear are negative in terms of sentiment, they can generate different political behavior.

Our dataset comes from \cite{widmann2022creating}. The corpus contains crowd-coded snippets with political content in German from three countries (Germany, Austria, and Switzerland) and two different sources (parliamentary speeches and Facebook posts).

In their study, the authors investigate eight emotions and compare several methods for emotion detection: An emotion dictionary, a classifier based on word embeddings, and a language model (ELECTRA). The strongest results are reported for 'Anger' (the most frequent emotion in the data), for which ELECTRA achieves an F1-Score of 0.84 (word embeddings = 0.79, dictionary = 0.59).

In our case study, we focus on \textit{Anger} detection. We perform some pre-processing to aggregate labels if multiple coders rated the same snippet. Our final dataset has $N = 7969$ observations with an imbalanced distribution of $2293$ angry (1) cases and $5676$ non-angry (0) cases.

Since the corpus is in German, we had two modeling options: Either (i) use the German text with a German-language LLM, or (ii) translate the corpus into English before fine-tuning our English-language models. We decided to evaluate both options. For the translation we used the DeepL API \footnote{https://www.deepl.com/en/pro-api}. Both versions are identical in terms of labels and class distribution.

\begingroup 
\renewcommand{\arraystretch}{1.2} 

\begin{table*}[h!]
\centering
\begin{threeparttable}
\caption{Results for Emotion Detection (Anger)}
\label{tab:exps-emo-angry}
\begin{tabular}{l l l l l l}
\hline
\textbf{Model Name} & \textbf{Accuracy} & \textbf{Prec. (wgt.)} & \textbf{Recall (wgt.)} & \textbf{F1 (macro)} & \textbf{F1 (wgt.)} \\
\hline
MAJ-VOT & $0.71\,\scriptstyle{(\pm0.00)}$ & $0.51\,\scriptstyle{(\pm0.00)}$ & $0.71\,\scriptstyle{(\pm0.00)}$ & $0.42\,\scriptstyle{(\pm0.00)}$ & $0.59\,\scriptstyle{(\pm0.00)}$ \\
\hline
ROB-BASE & $0.87\,\scriptstyle{(\pm0.01)}$ & $0.88\,\scriptstyle{(\pm0.01)}$ & $0.87\,\scriptstyle{(\pm0.01)}$ & $0.82\,\scriptstyle{(\pm0.01)}$ & $0.88\,\scriptstyle{(\pm0.01)}$ \\
ROB-LRG & $0.88\,\scriptstyle{(\pm0.01)}$ & $0.88\,\scriptstyle{(\pm0.00)}$ & $0.88\,\scriptstyle{(\pm0.01)}$ & $0.83\,\scriptstyle{(\pm0.00)}$ & $0.88\,\scriptstyle{(\pm0.00)}$ \\
DEB-V3 & $0.88\,\scriptstyle{(\pm0.01)}$ & $0.88\,\scriptstyle{(\pm0.00)}$ & $0.88\,\scriptstyle{(\pm0.01)}$ & $0.83\,\scriptstyle{(\pm0.01)}$ & $0.88\,\scriptstyle{(\pm0.00)}$ \\
ELE-LRG & $0.88\,\scriptstyle{(\pm0.00)}$ & $0.88\,\scriptstyle{(\pm0.02)}$ & $0.88\,\scriptstyle{(\pm0.00)}$ & $0.84\,\scriptstyle{(\pm0.00)}$ & $0.88\,\scriptstyle{(\pm0.00)}$\\
\textbf{XLNET-LRG} & $\bm{0.89\,\scriptstyle{(\pm0.00)}}$ & $\bm{0.89\,\scriptstyle{(\pm0.00)}}$ & $\bm{0.89\,\scriptstyle{(\pm0.00)}}$ & $\bm{0.85\,\scriptstyle{(\pm0.00)}}$ & $\bm{0.89\,\scriptstyle{(\pm0.00)}}$ \\
\hline
ELE-BS-GER & $0.88\,\scriptstyle{(\pm0.01)}$ & $0.88\,\scriptstyle{(\pm0.01)}$ & $0.88\,\scriptstyle{(\pm0.01)}$ & $0.83\,\scriptstyle{(\pm0.02)}$ & $0.88\,\scriptstyle{(\pm0.01)}$ \\
\midrule
BART-LRG & $0.26\,\scriptstyle{(\pm0.00)}$ & $0.36\,\scriptstyle{(\pm0.00)}$ & $0.26\,\scriptstyle{(\pm0.00)}$ & $0.24\,\scriptstyle{(\pm0.00)}$ & $0.29\,\scriptstyle{(\pm0.00)}$ \\
GPT-3.5 & $0.15\,\scriptstyle{(\pm0.00)}$ & $0.23\,\scriptstyle{(\pm0.00)}$ & $0.15\,\scriptstyle{(\pm0.00)}$ & $0.15\,\scriptstyle{(\pm0.00)}$ & $0.16\,\scriptstyle{(\pm0.00)}$ \\
GPT-4 & $0.20\,\scriptstyle{(\pm0.00)}$ & $0.18\,\scriptstyle{(\pm0.00)}$ & $0.20\,\scriptstyle{(\pm0.00)}$ & $0.18\,\scriptstyle{(\pm0.00)}$ & $0.13\,\scriptstyle{(\pm0.00)}$ \\
CLD-OPUS & $0.15\,\scriptstyle{(\pm0.00)}$ & $0.16\,\scriptstyle{(\pm0.00)}$ & $0.15\,\scriptstyle{(\pm0.00)}$ & $0.14\,\scriptstyle{(\pm0.00)}$ & $0.11\,\scriptstyle{(\pm0.00)}$ \\
\hline
\end{tabular}
\begin{tablenotes}[para,flushleft] \small
\textit{Note}: Results for fine-tuned models on unseen test set with $N=200$. Results for BART, GPTs, and Claude on full data. Fine-tuned models use gradient accumulation with 8 steps and batch size 4, except DEB-V3 (batch size 2). 
\end{tablenotes}
 \end{threeparttable}
\end{table*}

\endgroup

The prediction task is to identify whether a snippet contains an expression of anger. The results are presented in Table \ref{tab:exps-emo-angry}. XLNET-LRG achieves the best results, closely followed by ELECTRA, ROB-LRG, and DEB-V3. The ELECTRA model performs almost identically on the original German text and on the English translation. Translation can thus be a viable option when no language-specific pre-trained model exists.

BART, ChatGPT, and Claude all have considerable difficulties with this task and do substantially worse than our naive classifier. These results suggest that zero-shot classification works best for standard tasks but does not travel well to more specialized use-cases and nuanced semantics (which the models will have encountered less frequently during training).

\subsection{Multi-Class Stance Classification on Parties' EU Positions}

In our final study, we investigate European nationalist party positions. The dataset comes from \cite{Lfp2023}. It contains hand-coded evaluations of parties' positions toward the EU and European integration as reported in major European newspapers between the run-up to the 2016 Brexit referendum and the UK's 2020 EU exit. Positions range from acceptance of the European status-quo to outright demands to leave the EU.\footnote{The corpus covers ten EU countries. For a list of countries and parties, see \cite{Lfp2023}, Table 1).}

This setup implies a one-sided multi-class stance classification task: Stance because we consider party opposition (= opposing stance) toward European integration, one-sided because positions only range from neutral to strongly opposed (neglecting the supportive side of the spectrum), and multi-class because party positions are measured in three grades. 

\begingroup 
\renewcommand{\arraystretch}{1.2} 

\begin{table*}[h!]
\centering
\begin{threeparttable}
\caption{Results for Multi-Class Stance Classification (EU Positions)}
\label{tab:exps-brexit}
\begin{tabular}{l l l l l l}
\hline
\textbf{Model Name} & \textbf{Accuracy} & \textbf{Prec. (wgt.)} & \textbf{Recall (wgt.)} & \textbf{F1 (macro)} & \textbf{F1 (wgt.)} \\
\hline
MAJ-VOT & $0.83\,\scriptstyle{(\pm0.00)}$ & $0.68\,\scriptstyle{(\pm0.00)}$ & $0.83\,\scriptstyle{(\pm0.00)}$ & $0.30\,\scriptstyle{(\pm0.00)}$ & $0.75\,\scriptstyle{(\pm0.00)}$ \\
\hline
ROB-BASE & $0.84\,\scriptstyle{(\pm0.00)}$ & $0.87\,\scriptstyle{(\pm0.01)}$ & $0.84\,\scriptstyle{(\pm0.00)}$ & $0.70\,\scriptstyle{(\pm0.02)}$ & $0.85\,\scriptstyle{(\pm0.00)}$\\
ROB-LRG & $0.88\,\scriptstyle{(\pm0.01)}$ & $0.88\,\scriptstyle{(\pm0.01)}$ & $0.88\,\scriptstyle{(\pm0.01)}$ & $0.72\,\scriptstyle{(\pm0.03)}$ & $0.87\,\scriptstyle{(\pm0.01)}$ \\
\textbf{DEB-V3} & $\bm{0.92\,\scriptstyle{(\pm0.01)}}$ & $\bm{0.91\,\scriptstyle{(\pm0.01)}}$ & $\bm{0.92\,\scriptstyle{(\pm0.01)}}$ & $\bm{0.82\,\scriptstyle{(\pm0.02)}}$ & $\bm{0.91\,\scriptstyle{(\pm0.01)}}$ \\
ELE-LRG & $0.88\,\scriptstyle{(\pm0.01)}$ & $0.88\,\scriptstyle{(\pm0.01)}$ & $0.88\,\scriptstyle{(\pm0.01)}$ & $0.75\,\scriptstyle{(\pm0.03)}$ & $0.87\,\scriptstyle{(\pm0.01)}$ \\
XLNET-LRG & $0.87\,\scriptstyle{(\pm0.01)}$ & $0.89\,\scriptstyle{(\pm0.01)}$ & $0.87\,\scriptstyle{(\pm0.01)}$ & $0.75\,\scriptstyle{(\pm0.02)}$ & $0.88\,\scriptstyle{(\pm0.01)}$ \\
\hline
BART-LRG & $0.82\,\scriptstyle{(\pm0.00)}$ & $0.77\,\scriptstyle{(\pm0.00)}$ & $0.82\,\scriptstyle{(\pm0.00)}$ & $0.34\,\scriptstyle{(\pm0.00)}$ & $0.75\,\scriptstyle{(\pm0.00)}$ \\
GPT-3.5 & $0.24\,\scriptstyle{(\pm0.00)}$ & $0.65\,\scriptstyle{(\pm0.00)}$ & $0.24\,\scriptstyle{(\pm0.00)}$ & $0.17\,\scriptstyle{(\pm0.00)}$ & $0.27\,\scriptstyle{(\pm0.00)}$ \\
GPT-4 & $0.38\,\scriptstyle{(\pm0.00)}$ & $0.73\,\scriptstyle{(\pm0.00)}$ & $0.38\,\scriptstyle{(\pm0.00)}$ & $0.26\,\scriptstyle{(\pm0.00)}$ & $0.45\,\scriptstyle{(\pm0.00)}$ \\
CLD-OPUS & $0.26\,\scriptstyle{(\pm0.00)}$ & $0.75\,\scriptstyle{(\pm0.00)}$ & $0.26\,\scriptstyle{(\pm0.00)}$ & $0.25\,\scriptstyle{(\pm0.00)}$ & $0.29\,\scriptstyle{(\pm0.00)}$ \\
\hline
\end{tabular}
\begin{tablenotes}[para,flushleft] \small
\textit{Note}: Results for fine-tuned models on unseen test set with $N=200$. Results for BART, GPTs, and Claude on full data. Fine-tuned models use gradient accumulation with 8 steps and batch size 4, except DEB-V3 (batch size 2). 
\end{tablenotes}
 \end{threeparttable}
\end{table*}

\endgroup

We translate all news texts from the original languages to English using DeepL. We also aggregate the hand-coded euro-sceptic labels into three slightly broader categories: Status-quo acceptant or neutral statements are coded as 0 (least opposed), policy-level critique and reform demands are coded as 1, and institutional-level critique and outright leave-demands are coded as 2 (most opposed). We aggregate identical snippets by majority voting and ignore snippets coded as 'not relevant.'

The task is to predict neutral, moderately opposed, or strongly opposed positions toward the EU. The dataset has $N=3349$ observations and considerable class imbalance with 83\% of snippets in the moderately opposed category (neutral = $292$, moderate opposition, $2764$, strong opposition $293$). Results are shown in Table \ref{tab:exps-brexit}. DEB-V3 outperforms all other models, scoring well above 0.90 for most metrics. Most other fine-tuned models perform at least decently. BART, ChatGPT, and Claude, however, again do not exceed the naive baseline classifier — suggesting once more that the fine-tuning approach is superior when tasks are non-standard or context-specific.

\subsection{Fine-Tuning: The Effect of Training Set Size on Model Performance}

In this section, we turn to a complementary analysis aimed at understanding the training data requirements for the fine-tuning approach and how they affect the final model performance on the unseen test set. Among the five fine-tuned LLMs analyzed above, no single model outperforms the others on all tasks, but ROB-LRG and DEB-V3 are slightly ahead of the rest. Since ROB-LRG is less computationally demanding than DEB-V3, we consider it to provide the best overall balance.

For this analysis, we therefore revisit our four case studies with ROB-LRG, but evaluate model performance repeatedly at different training set sizes. For each run, we sample $N$ observations, where $N$ takes on values $N=\{50, 100, 200, 500, 1000\}$. The test set size remains at $N=200$ across all runs.

\begin{figure*}[h!]
    \centering
    \begin{minipage}[t]{0.48\textwidth}
    \includegraphics[width=\textwidth]{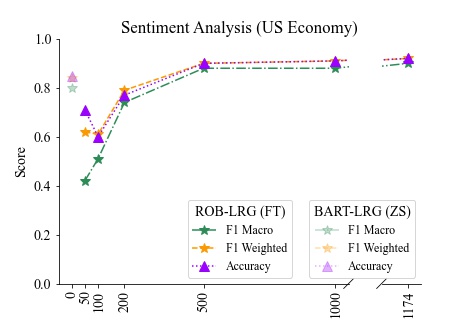}
    \end{minipage}
    \hfill
    \begin{minipage}[t]{0.48\textwidth}
   	\includegraphics[width=\textwidth]{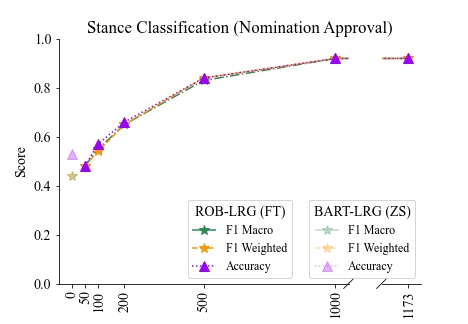}
    \end{minipage}
    \begin{minipage}[t]{0.48\textwidth}
    \includegraphics[width=\textwidth]{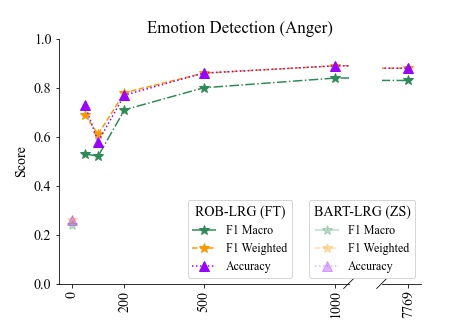}
    \end{minipage}
    \hfill
    \begin{minipage}[t]{0.48\textwidth}
    \includegraphics[width=\textwidth]{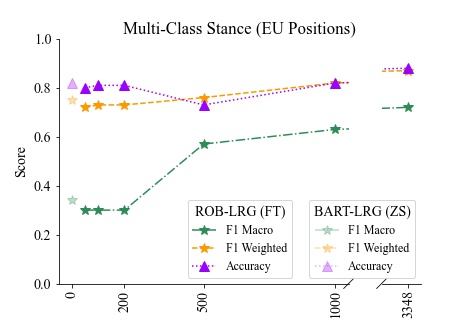}
    \end{minipage}
    \caption{Effect of training set size on model performance: Results for ROB-LRG with varying number of training observations $N=\{50, 100, 200, 500, 1000\}$. The translucent markers above the 0-point denote the zero-shot results of BART. The rightmost points denote model performance if trained on the full dataset.}
    \label{fig:ablation-dataset-size}
\end{figure*}

The results are depicted in Figure \ref{fig:ablation-dataset-size} and show how model performance increases with training data size. For model performance, we report F1 Macro, F1 Weighted, and Accuracy on the y-axis, evaluated on the unseen test set. For dataset size, we report sample size $N$ on the x-axis.

Overall, we see that performance picks up quickly as the amount of training data grows from minimal to moderate levels and then saturates as the training set size increases further.\footnote{See \cite{laurer2023less} for an insightful comparison of training efficiency between different machine learning approaches for text analysis.} This effect is most pronounced for more balanced datasets (e.g., the Kavanaugh Nomination Tweets). For imbalanced datasets (especially, EU positions data), we see high (but meaningless) accuracy scores at low sample sizes because poorly trained models always predict the majority class. Macro-F1 scores, however, which punish naive majority class prediction, consistently show the saturation pattern. Overall, the sweet-spot tends to lie between 200 and 500 training observations. Below 200 observations, model performance remains below its potential. Above 500 observations, performance begins to level off.

We also show the results of BART, arguably the most consistent of the zero-shot approaches overall (compared to GPT-3.5, GPT-4, and Claude Opus). Since the training set for zero-shot classification is essentially zero, we show the BART results on the left of each plot (x-axis value = $0$) with slightly translucent markers. As can be seen, ROB-LRG begins to outperform BART on all datasets when the training set exceeds $N=200$.

\section{Discussion}

Our key finding is that fine-tuning smaller LLMs with dedicated training data is consistently superior to zero-shot prompting larger models such as BART, ChatGPT, and Claude Opus. While zero-shot results are decent for common tasks such as sentiment analysis, fine-tuned LLMs always surpass zero-shot performance as training set size increases. This tendency is more pronounced for less standard tasks such as stance classification or emotion detection.

An interesting question is whether next-generation generative AI models will catch up with, or even surpass, fine-tuned LLMs in performing specialized tasks. In the following, we point to some areas where we expect progress regarding both fine-tuning and prompt-based approaches, which may be helpful in structuring expectations about future trajectories: (i) data augmentation, (ii) secondary pre-training, (iii) model architecture, and (iv) prompt engineering and few-shot learning. Lastly, we discuss some general considerations when choosing between smaller BERT-like LLMs and larger generative AI models in production settings.  

\begin{itemize}

    \item \textit{Data Augmentation}: Data augmentation is an approach to synthetically increase labeled training data without human input. One approach in NLP is \textit{back translation}, where the original text is translated into another language and then back-translated. Adding the back-translated sentence to the training data (with the same label as the original) can boost performance as shown by \cite{sennrich2015improving}. Another option is \textit{token perturbation}, where certain words are replaced with synonyms \cite{wei2019eda}. Going forward, these techniques can further reduce data requirements for fine-tuning LLMs.

    \item \textit{Secondary Pre-training}: An additional pre-training phase on a smaller domain-specific corpus can improve both fine-tuned encoder-style and instruction-tuned decoder-style LLMs. For RoBERTa, \cite{gururangan2020don} demonstrated significant performance gains for downstream classification tasks when further pre-trained in the specific domain to gain more knowledge about a specific vocabulary and usage before performing downstream fine-tuning for the classification task.

    \item \textit{Model Architecture}: While there has been a trend toward larger and larger models, recent work has shown that scaling from hundreds of billions to trillions of parameters yields saturating improvements and that scale alone may not be the single factor that determines performance \cite{schaeffer2023emergent, caballero2022broken}. Instead, research has been devoted to investigate \textit{multi-modality} — the integration of multiple modes of information such as text, image and audio \cite{driess2023palm} —,'ensemble' (or mixture-of-experts) architectures \cite{shazeer2017outrageously}, and ways to improve training data quality while keeping model size in the billion parameter range \cite{gunasekar2023textbooks}.

    \item \textit{Prompt Engineering and Few-Shot Learning}: Prompt engineering can boost the performance of instruction-tuned LLMs such as ChatGPT and Claude. One option is \textit{few-shot learning}, which results in so-called \textit{in-context} learning by providing labeled training data as part of the prompt. For GPT-3.5, few-shot prompting yielded some performance gains as seen in \cite{brown2020language}\footnote{In exploratory experiments on our datasets, few-shot prompting did not improve results for GPT-3.5 and 4.0.}. However, few-shot learning requires careful selection of examples to provide in the prompt and systematic evaluation of prompt designs. This significantly increases the required time investment and multiplies costs due to the increased length and number of prompts. Consequently, few-shot learning risks negating the simplicity and intuitive handling advantages of generative AI models. Nonetheless, with further developments in model architecture, in-context learning via few-shot prompting may eventually surpass fine-tuned encoder-style LLMs.

\end{itemize}

Depending on how the above developments play out, the relative classification performance of smaller BERT-style LLMs and larger generative AI models may shift to one side or the other. However, at present, fine-tuning of smaller models remains the superior approach for text classification. 

Apart from performance, we see additional advantages of relying on smaller, fine-tuned LLMs in production settings: 

With an eye to trade secrets or privacy, smaller models provide an unmatched level of control. Because these models are relatively small, they can be stored, trained, and run on a local machine. Relatedly, all model parameters are visible and can be saved and backed up, providing programmatic stability and predictability. This is in stark contrast to proprietary, cloud-based generative AI models, which require information to be transferred to an external black-box model.    

Additionally, it is important to remember that professional use of generative AI models requires labeled data for validation — even when using zero-shot prompting. Given the relatively better performance of smaller fine-tuned models with limited training data, along with their advantages regarding control and privacy, it seems likely that models like BERT and similar LLMs will remain valuable tools for years to come.

\section{Conclusion}

In this paper, we investigate whether zero-shot prompted generative AI models such as ChatGPT and Claude Opus can already outperform fitted models such as smaller fine-tuned LLMs for text classification tasks. To this end, we compare several fine-tuned LLMs such as RoBERTa with major generative AI models across text categories, tasks, languages, and imbalanced datasets. 

Our results show that fine-tuning smaller LLMs still significantly outperforms zero-shot prompted generative AI models on all our case studies. In addition, we provide an ablation study to explore the relationship between dataset size (i.e., number of training examples) and model performance, showing that performance already begins to saturate after around 200 labels. 

We further provide an easy-to-use text classification toolkit that requires no domain knowledge in NLP or deep learning. Our toolkit is modular and allows users to plug in additional models as desired or needed. This allows users to load smaller LLMs for other languages or upgrade to newly developed or improved language models and as they are published via Hugging Face. We hope that our tool will enable users to realize the potential of fine-tuned LLMs.

\paragraph{Acknowledgments} 
The authors would like to thank Daniel Grosshans, Sascha Langenbach, Janina Steinmetz, and Stefanie Walter.

\paragraph{Funding Statement} This project has received funding from the European Research Council (ERC) under the European Union’s Horizon 2020 research and innovation programme grant agreement No 817582 (ERC Consolidator Grant DISINTEGRATION).

\paragraph{Replication materials} All code and replication materials are available \href{https://github.com/mnbucher/text-cls-llms}{here}.

\printbibliography

\newpage
\section*{Appendix}
\section*{A: Zero-Shot Prompts for GPT-3.5, GPT-4, and Claude Opus}
After experimenting through the API with a few prompts, we fine-tuned our final prompt to maximize the likelihood that the model only returns one label from the given list of available class labels. As LLMs are discrete probabilistic models, it's stochasticity can lead to a variance in its generated output. By controlling the temperature parameter and decreasing its value, we can decrease the variance of the generated output. However, the model still generates other content than just one of the given output labels. In the following, we present the prompts we provided to the generative AI APIs, where <Text> marks the text placeholder for the current text sample. We set \texttt{temperature=0.1} for all runs to reduce variance and force the model to produce a single label output from the provided list of labels. We repeat the prompting if the output does not match the labels provided for the classification task until the model produces a correct label as output.

\subsection*{Sentiment Analysis on The New York Times Coverage of the US Economy}

Prompt:

\blockquote{
You have been assigned the task of zero-shot text classification for sentiment analysis. Your objective is to classify a given text snippet into one of several possible class labels, based on the sentiment expressed in the text. Your output should consist of a single class label that best matches the sentiment expressed in the text. Your output should consist of a single class label that best matches the given text. Choose ONLY from the given class labels below and ONLY output the label without any other characters.

Text: <Text>

Labels: 'Negative Sentiment', 'Positive Sentiment'

Answer:
}

\subsection*{Stance Classification on Tweets about Kavanaugh Nomination}

Prompt:

\blockquote{
You have been assigned the task of zero-shot text classification for stance classification. Your objective is to classify a given text snippet into one of several possible class labels, based on the attitudinal stance towards the given text. Your output should consist of a single class label that best matches the stance expressed in the text. Your output should consist of a single class label that best matches the given text. Choose ONLY from the given class labels below and ONLY output the label without any other characters.

Text: <Text>

Labels: 'negative attitudinal stance towards', 'positive attitudinal stance towards'

Answer:
}

\subsection*{Emotion Detection on Political Texts in German}

Prompt:

\blockquote{
You have been assigned the task of zero-shot text classification for emotion classification. Your objective is to classify a given text snippet into one of several possible class labels, based on the anger level in the given text. Your output should consist of a single class label that best matches the anger expressed in the text. Choose ONLY from the given class labels below and ONLY output the label without any other characters.

Text: <Text>

Labels: 'Angry', 'Non-Angry'

Answer:
}

\subsection*{Multi-Class Stance Classification on Parties’ EU Positions}

Prompt:

\blockquote{
You have been assigned the task of zero-shot text classification for political texts on attitudinal stance towards Brexit and leave demands related to the European Union (EU). Your objective is to classify a given text snippet into one of several possible class labels, based on the stance towards Brexit and general leave demands in the given text. Your output should consist of a single class label that best matches the content expressed in the text. Choose ONLY from the given class labels below and ONLY output the label without any other characters.

Text: <Text>

Labels: 'Neutral towards Leave demands', 'Pro-Leave demands', 'Very Pro-Leave demands'

Answer:
}

\section*{B: BART}
For BART, we use the BART Large model \texttt{facebook/bart-large-mnli} introduced by \cite{lewis2019bart} together with the Zero-Shot Classification pipeline from Huggingface. We use a technique called Natural Language Inference (NLI), which prompts a model using two sentences, a \textit{Premise} (in our case the text to be classified) and a \textit{Hypothesis} (a possible class label for that
text). The model then predicts if the hypothesis is consistent with the premise. NLI evaluates a hypothesis for each label and then selects the label with the highest confidence as output.

\end{document}